\documentclass{article}

\usepackage{microtype}
\usepackage{graphicx}
\usepackage{subfigure}
\usepackage{booktabs} 

\usepackage{hyperref}

\usepackage[accepted]{icml2025}

\usepackage{amsmath}
\usepackage{amssymb}
\usepackage{mathtools}
\usepackage{amsthm}
\usepackage{enumitem}
\usepackage{pifont}
\usepackage{inconsolata}
\usepackage{xcolor}
\newcommand{\cmark}{\textcolor{green!60!black}{\ding{51}}}%
\newcommand{\xmark}{\textcolor{red}{\ding{55}}}%

\makeatletter
\newcommand{\subalign}[1]{%
  \vcenter{%
    \Let@ \restore@math@cr \default@tag
    \baselineskip\fontdimen10 \scriptfont\tw@
    \advance\baselineskip\fontdimen12 \scriptfont\tw@
    \lineskip\thr@@\fontdimen8 \scriptfont\thr@@
    \lineskiplimit\lineskip
    \ialign{\hfil$\m@th\scriptstyle##$&$\m@th\scriptstyle{}##$\hfil\crcr
      #1\crcr
    }%
  }%
}
\makeatother

\usepackage[capitalize,noabbrev]{cleveref}

\theoremstyle{plain}

\theoremstyle{definition}

\theoremstyle{remark}

\usepackage[textsize=tiny]{todonotes}

\icmltitlerunning{Submission and Formatting Instructions for ICML 2025}

\begin{document}

\twocolumn[
\icmltitle{Enabling Autoregressive Models to Fill In Masked Tokens }



\icmlsetsymbol{equal}{*}

\begin{icmlauthorlist}
\icmlauthor{Daniel Israel}{ucla}
\icmlauthor{Aditya Grover}{ucla}
\icmlauthor{Guy Van den Broeck}{ucla}
\end{icmlauthorlist}

\icmlaffiliation{ucla}{Department of Computer Science, University of California Los Angeles, Los Angeles, USA}

\icmlcorrespondingauthor{Daniel Israel}{disrael@cs.ucla.edu}
\icmlcorrespondingauthor{Aditya Grover}{adityag@cs.ucla.edu}
\icmlcorrespondingauthor{Guy Van den Broeck}{guyvdb@cs.ucla.edu}

\icmlkeywords{Machine Learning, ICML}

\vskip 0.3in
]



\printAffiliationsAndNotice{}  


\begin{abstract}

Historically, LLMs have been trained using either autoregressive (AR) or masked language modeling (MLM) objectives, with AR models gaining dominance in recent years. However, AR models are inherently incapable of masked infilling, which is the ability to predict masked tokens between past and future context. In contrast, MLM models suffer from intrinsic computational inefficiencies during both training and inference that hinder their scalability. This work introduces MARIA (Masked and Autoregressive Infilling Architecture), a novel approach that leverages the strengths of both paradigms to achieve state-of-the-art masked infilling performance. MARIA combines a pre-trained MLM and AR model by training a linear decoder that takes their concatenated hidden states as input. This minimal modification enables the AR model to perform infilling while retaining its inherent advantages in terms of faster inference with KV caching. Our results demonstrate that MARIA significantly outperforms existing methods, namely discrete diffusion models, on masked infilling tasks.
\end{abstract}
\section{Introduction}


The field of natural language processing (NLP) has witnessed remarkable advancements in recent years, largely driven by the advent of large language models (LLMs) \cite{zhao2023survey} built upon the Transformer architecture \cite{vaswani2017attention}. These models, characterized by their self-attention mechanisms and vast parameter counts, have demonstrated unprecedented capabilities in understanding and generating human-like text.

A critical aspect of LLM training lies in the choice of pre-training objective. Traditionally, two dominant paradigms have emerged: autoregressive (AR) and masked language modeling (MLM). AR models, such as GPT \cite{achiam2023gpt}, are trained to predict the next token in a sequence, given the preceding context. This left-to-right approach, coupled with causal masking that prevents the model from ``seeing" future tokens, enables efficient training and inference. MLM models, exemplified by BERT \cite{devlin2019bertpretrainingdeepbidirectional}, are trained to predict masked-out tokens in a sequence, leveraging bidirectional context from both past and future tokens.

One notable capability where AR models typically fall short is text infilling \cite{donahue2020enabling}, the task of predicting missing tokens within a given text span, surrounded by both preceding and subsequent context. While MLM models inherently support infilling due to their bidirectional nature, AR models, with their unidirectional processing, cannot leverage future context for this task.  This limitation restricts the applicability of AR models in scenarios where infilling is essential, such as interactive text editing \cite{lee2022coauthor}, code completion \cite{liu2020multi}, and structured generation \cite{xia2024fofo}. 

\begin{table*}[t]
\label{tab:maria_summary}
    \centering
\begin{tabular}{l c c c}
\toprule
 Model & Scalable Training & KV Cached Inference  & Supports Mask Infilling   \\
 \midrule
 AR & \cmark & \cmark & \xmark \\
 MLM & \xmark & \xmark &  \cmark \\
 MARIA & \cmark & \cmark & \cmark \\

\bottomrule
\end{tabular}
\caption{\textbf{Comparison of different modeling approaches.} We compare the three modelling approaches: Autoregressive (AR), Masked Language Modelling, and our method Masked and Autoregressive Infilling Architecture (MARIA). While AR enjoys more scalable training and computationally efficient inference, it cannot perform masked infilling. Contrarily, MLM can but is less scalable. We argue that our method MARIA inherits the benefits from both approaches.}
\end{table*}

Despite the limitations of AR models in handling text infilling, they remain the dominant paradigm for large-scale language modeling due to their superior scalability. AR models benefit from several key advantages that make them more efficient during both training and inference. First, AR models can exploit causal masking to parallelize every next token prediction, enabling faster training on massive datasets across multiple GPUs. This differs from MLM models, which only make predictions for a fixed ratio of masked tokens during training, such as 15 percent in BERT. Second, the sequential nature of AR models allows for the use of KV caching at inference time, which significantly reduces the computational cost of attention operations by reusing previously computed embeddings. Significant effort has been dedicated to optimizing the memory and speed of KV caching \cite{kwon2023efficient, zhao2024prepacking, liu2024scissorhands}. Thus, AR models are better suited for real-time applications, such as chatbots and virtual assistants, where low-latency responses are critical. These factors contribute to the widespread adoption of AR models in industry and academia, despite their inherent limitations for infilling.

Researchers have explored non-autoregressive paradigms that support text infilling. One such approach is discrete diffusion \cite{lou2023discrete}, which iteratively refines a noisy input sequence. Discrete diffusion models have shown promise in tasks like text generation and infilling. However, discrete diffusion models are built on the MLM modeling paradigm, making it difficult to scale their training in the same manner as AR models. Furthermore, these models often require numerous refinement steps and do not support KV caching, which can make them less efficient for inference. 

Given the complementary strengths and weaknesses of AR and MLM models, there is a clear need for a hybrid approach that leverages the best of both paradigms. In this work, we introduce MARIA (Masked and Autoregressive Infilling Architecture), a novel framework that combines the benefits of AR and MLM models to achieve state-of-the-art performance in text infilling. MARIA integrates a pre-trained MLM and AR model by training a linear decoder that takes the concatenated hidden states of both models as input. This minimal modification enables the AR model to perform effective infilling while retaining its inherent advantages in terms of faster inference with KV caching. Our experiments demonstrate that MARIA significantly outperforms existing methods, including discrete diffusion models, on a variety of text infilling benchmarks. By bridging the gap between AR and MLM paradigms, MARIA offers a new technique for scaling infilling language models. We summarize the advantages of MARIA in Table \ref{tab:maria_summary}.

\section{Related Works}
\subsection*{Discrete Diffusion}
Discrete diffusion models have emerged as a promising alternative to traditional autoregressive models for text generation and, notably, text infilling. Inspired by the success of diffusion models in continuous domains like image generation \cite{ho2020denoisingdiffusionprobabilisticmodels}, these models adapt the diffusion framework to operate on discrete sequences of tokens. In the context of text infilling, discrete diffusion offers several advantages. Its iterative refinement process allows for fine-grained control over the generated text and the ability to tradeoff quality for efficiency. However, as mentioned in the introduction, these models can be computationally expensive during inference due to the multiple refinement steps and the lack of KV caching.  They also face challenges in scaling up training compared to autoregressive models. In this paper, we will primarily focus on the work of Scaling Masked Diffusion Model (SMDM) \cite{nie2024scalingmaskeddiffusionmodels} and DiffuLlama \cite{gong2024scalingdiffusionlanguagemodels}, but the space includes many promising works \cite{sahoo2024simpleeffectivemaskeddiffusion, liu2024discrete, liu2024thinkgeneratediscretediffusion, hoogeboomardiffusion, ou2024absorbingdiscretediffusionsecretly}
\subsection*{FIM}
AR models can be adapted to perform infilling through a special training process called Fill-in-the-Middle (FIM) \cite{donahue2020enabling}, in which the order of the original sequence is changed such that the middle of the sequence is moved to the end and marked with a special FIM token. These FIM models are particularly useful for coding applications \cite{fried2023incodergenerativemodelcode}. We make a distinction between FIM and masked infilling. FIM necessitates that the infilled text is a contiguous block, while masked infilling can fill in arbitrary sequences of tokens.
\subsection*{MLM and AR Unification}
Notable works to unify MLM and AR modelling include BART \cite{lewis2019bart}. Besides architectural differences, the main distinction between MARIA and BART is that MARIA is applied to existing pretrained MLM and AR models, while BART must be trained end-to-end. Other notable works incorporate together MLM and AR modeling techniques for improved training \cite{du2022glmgenerallanguagemodel, nguyen2023meetmiddlenewpretraining, yu2024antlmbridgingcausalmasked}, but none are expressly targeting masked infilling as an application.
\section{Method}
\label{sec:method}
\begin{figure*}
    \centering
    \includegraphics[width=0.8\linewidth]{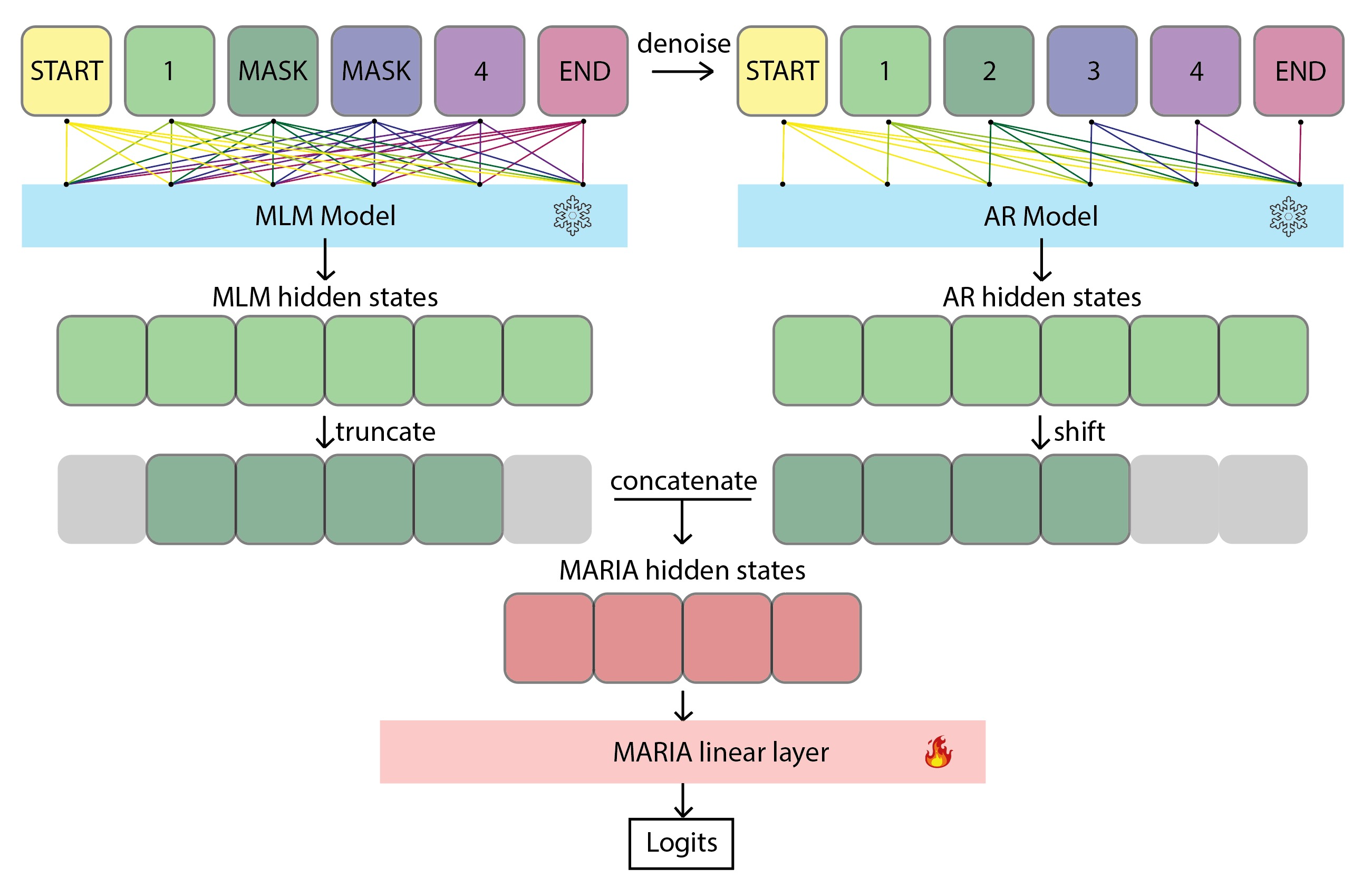}
    \caption{\textbf{MARIA architecture and training pipeline.} MARIA takes two frozen pretained models: one MLM and one AR. As input, the MLM recieves the masked inputs and the AR model recieves the denoised inputs. We compute the hidden states under each model and perform truncating and shifting operations to ensure both hidden states model the same tokens. MARIA trains a linear layer to predict the logits of each masked input on the concatenated hidden states. This training scheme models an autoregressive distribution conditioned on unmasked tokens.}
    \label{fig:maria_diagram}
\end{figure*}
\section*{Background}
Consider an autoregressive model $\pi_{\mathrm{AR}}$ and masked language model $\pi_{\mathrm{MLM}}$. Given access to a dataset $\mathcal{D} = \{x_1, x_2, ...\}$, autoregressive models are trained to maximize the joint likelihood given by


\begin{equation}
    \mathcal{L}_{\mathrm{AR}} = -\mathbb{E}_{x\sim \mathcal{D}} \left[ \sum_{i} \log \pi_{\mathrm{AR}}(x_i \mid x_{<i}) \right]
\end{equation}
Masked language models employ a masking objective  that assumes a distribution over masks $\mathcal{M}$, where $m \in \mathcal{M}$ is selection of indices $m = \{i_1, i_2, ...\}$.
\begin{equation}
    \mathcal{L}_{\mathrm{MLM}} = -\mathbb{E}_{\substack{x \sim \mathcal{D}\\m \sim \mathcal{M}}}\left[ \sum_{i \in m} \log \pi_{\mathrm{MLM}}(x_i \mid x_{\setminus m}) \right]
\end{equation}
Also observe that each language model is composed of a function $h$ that embeds inputs into hidden state vectors and a linear weight matrix $W$ used to decode the hidden states into logits.
\begin{equation}
    \pi_{\mathrm{AR}}(x \mid \cdot\,) = \sigma\left(W_1 h_1(x)\right) 
\end{equation}
\begin{equation}
    \pi_{\mathrm{MLM}}(x \mid \cdot\,) = \sigma\left(W_2 h_2(x)\right)
\end{equation}
where $\sigma(z_i) = e^{z_i}/ \Sigma_j e^{z_j}$ is the softmax function. We define $W_1 \in \mathbb{R}^{d_1 \times v}$ and $W_2 \in \mathbb{R}^{d_2 \times v}$ such that their hidden dimensions $d$ can be different but vocabulary size $v$ are the same.
\section*{MARIA}
\subsection*{Objective}
The MARIA architecture can be defined very straightforwardly with a linear layer on the concatenated hidden states of an AR and MLM model.
\begin{equation}
    \pi_{\mathrm{MARIA}}(x \mid \cdot\,) = \sigma(W_3 \, [h_1(x); h_2(x)])
\end{equation}
where $W_3 \in \mathbb{R}^{(d_1+d_2) \times v}$. Finally, we may now define an objective that is both autoregressive and masked. Let $c(i, m) = \{x_{<i}, x_{>i \, \cap \, \setminus m }\}$ define the union of tokens before the index $i$  and all unmasked tokens after $i$.
\begin{equation*}
\mathcal{L}_{\mathrm{MARIA}} = -\mathbb{E}_{\substack{x \sim \mathcal{D}\\m \sim \mathcal{M}}}\left[ \sum_{i \in m} \log \pi_{\mathrm{MARIA}}(x_i \mid c(i, m) \right]
\end{equation*}
This objective defines the expected negative log likelihood of an autoregressive distribution conditioned on unmasked tokens.
\subsection*{Training Procedure}
MARIA training can be parallelized in a similar manner as a typical autoregressive Transformer. For a clean input sequence $X_{1:n}$, we consider its masked counter part $M_{1:n}$. The AR model receives the clean inputs and the MLM model receives the masked inputs such that we compute the hidden state $[h_1(X); h_2(M)]_{1:n}$ concatenated on the sequence dimension. These are then decoded with $W_3$ to next token logits. Thus, the autoregressive loss is computed over the entire sequence in parallel. This training procedure is best depicted by Figure \ref{fig:maria_diagram}. 
\subsection*{Initialization}
\begin{figure}
    \centering
    \includegraphics[width=1\linewidth]{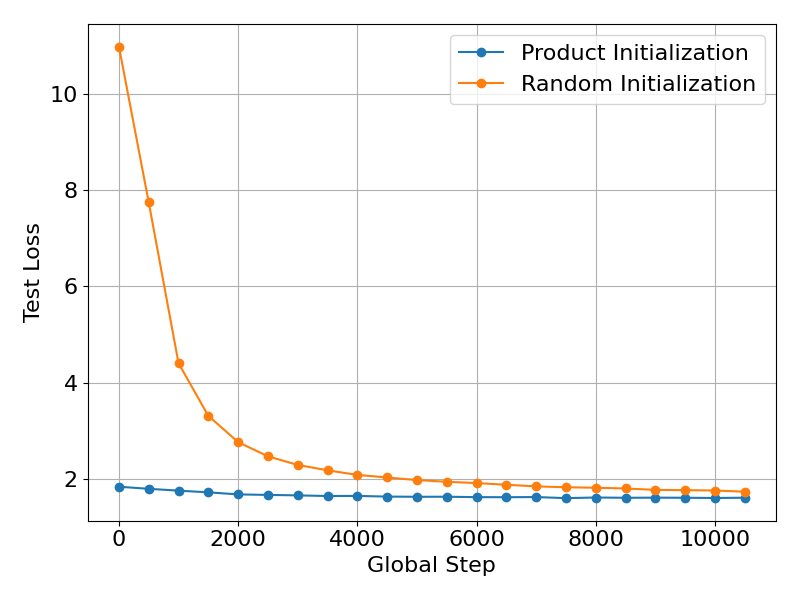}
    \caption{\textbf{Comparing evaluation loss curves for two different weight initializations.} Product initialization (ours) is a far better weight initialization than random weight initialization, leading to faster training and better convergence.}
    \label{fig:init}
\end{figure}

As part of our method, we also provide a way to initialize the newly defined MARIA weights $W_3$. Because we have access to existing weights of pretrained models, namely an autoregressive weights $W_1$ and masked weights $W_2$, we can initialize $W_3$
\begin{equation}
    W_3 \leftarrow [W_1/2; W_2/2]
\end{equation}
Observe that this will output the average of the logits of $\pi_{\mathrm{AR}}$ and $\pi_{\mathrm{MLM}}$
\begin{align}
    \pi_{\mathrm{MARIA}}(x \mid \cdot\,) &= \sigma([W_1/2; W_2 / 2] \, [h_1(x); h_2(x)]) \\
    &= \sigma((\pi_{\mathrm{AR}}(x \mid \cdot\,)+\pi_{\mathrm{MLM}}(x \mid \cdot\,))/2)
\end{align}
This is a good initialization because the average of logits corresponds to a multiplicative mixture of the two original distributions. This ensemble, known as product of experts \cite{hinton2002training}, has proven effective in the context of LLMs \cite{liu2021dexperts}. Smart weight initialization leads to faster and better convergence \cite{samragh2024scaling}, and we demonstrate this with ``product initialization" for MARIA in Figure \ref{fig:init}.
\subsection*{Unconditional Generative Model}
While MARIA is trained to sample conditionally, we propose a method to sample unconditionally. The desideratum of this generative model is to make possible iterative refinement of text such that more compute leads to better samples. Discrete diffusion has this property for the number of denoising steps, and while it is possible to use MARIA directly as a discrete diffusion model, it is undesirable because autoregressive sampling at each times step is slow, and discrete diffusion only unmasks a small number of tokens at a time, remasking most samples at every iteration. Thus, we propose using MARIA as a generative model with an inference strategy inspired by simulated annealing \cite{bertsimas1993simulated}. We can describe the process as follows:
\begin{enumerate}
    \item Sample from the base AR model at temperature 1.
    \item Using MARIA, resample a fixed percentage of tokens autoregressively at temperature $T$.
    \item Repeat the process for some number of iterations, annealing $T$ from 1 to 0.
\end{enumerate}
This inference strategy is a way to optimize over the joint likelihood of a sequence, and it is an improvement over standard greedy sampling because it is non-myopic \cite{shih2023long}. More formally, we are sampling from the following distribution:
\begin{align*}
    &p\left(x^{i}\right)\propto\\
    &\sum_{\subalign{x^{1:i-1}\\m^{1:i-1}}}\prod_{j=1}^i  \pi_{\mathrm{MARIA}}\left(x^{j}\,\Big|\, x^{j-1},m^{j-1} ; t_{j-1} \,\right)
\end{align*}

where $x^{i}$ is the sequence at step $i$, $t_j$ is a temperature at step $j$, $m^{k}$ is a mask at step $k$, and $\pi_{\mathrm{MARIA}}(\,\cdot \,; t)$ denotes the autoregressive MARIA distribution temperature scaled by $t$.

\section*{Implementation}
\subsection*{Models}
A key constraint of MARIA is that the combined AR and MLM models must be trained with the same tokenizer. We make use of two important open-source works that are both trained with a GPT2 \cite{radford2019language} based tokenizer: ModernBERT \cite{warner2024smarter} and OLMo \cite{groeneveld2024olmo}. We train two models: 
\begin{itemize}
    \item MARIA 1B: a model composed of ModernBERT-Large and pretrained OLMo 1B
    \item MARIA 7B: a model composed of ModernBERT-Large and pretrained OLMo 7B
\end{itemize}
We will refer to these models in this manner throughout the course of the paper.


\subsection*{Training}
Our training data is composed of high quality tokens from FineWebEdu \cite{penedo2024finewebdatasetsdecantingweb}, a standard pretraining corpus curated for fast convergence and good downstream performance. We randomly mask the data by sampling masking rates from a Beta(2.5, 2.5) distribution, which is more effective than a uniform rate \cite{shen2023filmfillinlanguagemodels}.
To train the MARIA Linear Layer, we initialize the weights as previously described. For MARIA 1B and MARIA 7B respectively, we train for 90000 steps (approximately 30 billion tokens) and 25000 steps (approximately 7 billion tokens). Given the size of FineWebEdu, we complete less than a single epoch, and we evaluate test loss on ten thousand holdout examples. We train at batch size 32 using gradient accumulation with a learning rate of 5-e5 and cosine learning rate schedule. Our training hardware is comprised of 8 NVIDIA 48GB A6000 GPUs connected to a Colfax CX41060s-EK9 4U Rackmount Server with AMD EPYC (Genoa) 9124 processors.

\subsection*{Inference}

\begin{algorithm}[t]
\caption{MARIA KV Cached Inference}
\label{alg:maria_inference}
\begin{algorithmic}[1]
    \STATE \textbf{Input:} \texttt{input\_ids}, \texttt{masked\_indices}
    \STATE \textbf{Output:} \texttt{input\_ids} with infilled \texttt{[MASK]} tokens\\\COMMENT{Get MLM hidden states once}
    \STATE \texttt{mlm\_hidden\_states} $\gets$ \text{MLM\_Model}\texttt{(input\_ids)}
    \STATE \texttt{past\_kv} $\gets$ \texttt{None}
    \STATE \texttt{prev\_idx} $\gets$ 0
    \FOR{\texttt{curr\_idx} $\in$ \texttt{masked\_indices}}
        \STATE \texttt{ar\_input} $\gets$ \texttt{input\_ids[prev\_idx:curr\_idx]}\\\COMMENT{Run AR model with caching}
        \STATE \texttt{ar\_output} $\gets$ \texttt{\text{AR\_Model}(ar\_input, past\_kv)}\\\COMMENT{Update cache}
        \STATE \texttt{past\_kv} $\gets$ \texttt{ar\_output.\text{past\_kv}}
        \STATE \texttt{ar\_hidden\_state} $\gets$ \texttt{ar\_output.\text{hidden\_states}}
        \STATE \texttt{maria\_hidden\_states} $\gets$ \texttt{\text{Concat}(}\\ \qquad\texttt{ar\_hidden\_state,}\\ \qquad \texttt{mlm\_hidden\_states[curr\_idx]} \\)
        \STATE \texttt{logits} $\gets$ \texttt{\text{MARIA\_Linear}(maria\_hidden\_states)}
        \STATE \texttt{sampled\_token} $\gets$ \texttt{\text{Sample}(logits)}\\\COMMENT{Fill in the mask}
        \STATE \texttt{input\_ids[curr\_idx]} $\gets$ \texttt{sampled\_token} 
        \STATE \texttt{prev\_idx} $\gets$ \texttt{curr\_idx}
    \ENDFOR
    \STATE \textbf{return} \texttt{input\_ids}
\end{algorithmic}
\end{algorithm}

As we will further argue in Section \ref{sec:experiments}, AR models have an advantage at inference time over MLM models with the ability to reuse previous computations through KV caching. Transformers with bidirectional masking cannot cache the computations from previous samples because sampling a new token will change the representations of all existing future tokens. We present a simple KV caching inference algorithm with MARIA in Algorithm \ref{alg:maria_inference}. This algorithm computes a single forward pass on the MLM model to compute hidden states. After this negligible overhead, we perform standard KV caching just the same as a standard AR model.

\section{Experiments}

\begin{table*}[t]
    \centering
\begin{tabular}{l r r r r r r r}
 \toprule & \\[-2ex]
 Model & Size & Type  & Masking Rate & & & &  \\[0.5ex]
 \hline & \\[-2ex]
 & &  & 0.1 & 0.3 & 0.5 & 0.7 & 0.9 \\[0.5ex]
 \hline & \\[-2ex]

ModernBert & 0.395 B & MLM (AR Decode) & 2.92 & 5.79 & 19.73 & 136.2 & 1468 \\
OLMo 1B & 1.18 B & AR & 22.28 & 22.13 & 22.20 & 22.17 & 22.62 \\
OLMo 7B & 7.3 B & AR & 14.93 & 15.01 & 14.96 & 15.00 & 15.046 \\
SMDM &1.1B & DD & $\leq$ 14.44 & $\leq$ 46.36 & $\leq$ 118.7 & $\leq$ 363.7 & $\leq$ 1391 \\
DiffuLlama & 6.74 B & DD & $\leq$ 10.36 & $\leq$ 30.04 & $\leq$ 68.38 & $\leq$ 180.5 & $\leq$ 599.5\\[1ex]
\hline & \\[-1.5ex]
MARIA 1B (ours) & 1.575 B &MLM + AR& 3.10 & 4.45 & 7.41 & 13.80 & 23.99 \\
MARIA 7B (ours) & 7.695 B & MLM + AR & \textbf{2.82} & \textbf{3.85} & \textbf{5.94} & \textbf{10.11} & \textbf{16.30} \\[1ex]
\bottomrule & \\[-1.5ex]
\end{tabular}
\caption{\textbf{Downstream perplexity for various masking ratios.} We evaluate the downstream perplexity, averaging over 5 standard evaluation sets. ModernBERT is computed autoregressively, and we estimate the upper bound perplexity in the discrete diffusion models. MARIA performs the best by inheriting the strengths of its components: OLMo (AR) and ModernBERT (MLM). Based on parameter counts, MARIA presents the most effective way to scale models for masked token infilling.}
\end{table*}

\label{sec:experiments}
In this section, we evaluate MARIA in a variety of settings against strong baselines. Our key findings include:
\begin{itemize}
    \item \textbf{Superior Perplexity.} MARIA achieves lower perplexity across various masking rates and datasets compared to ModernBERT, SMDM, and DiffuLlama.
    \item \textbf{Efficient Inference.} MARIA offers high throughput by KV caching at inference time. AR decoding with ModernBERT does not scale.
    \item \textbf{High-Quality Samples.} Evaluation using LLM judge based ELO demonstrates that MARIA's generated text is of higher quality than baselines.
    \item \textbf{Better Representations.} MARIA exhibits better representations for a downstream part-of-speech tagging task.
\end{itemize}
\section*{Baselines}
We consider three primary baselines to compare our method against. First, we consider ModernBERT. Although ModernBERT is an MLM model, in practice MLM models can be used autoregressively by progressively filling in masks from left to right. Surprisingly, MLM models demonstrate considerable in-context learning capabilities when used in this manner \cite{samuel2024bertsgenerativeincontextlearners}. Another necessary baseline for masked infilling are discrete diffusion models, of which we select Scaling Masked Diffusion Model (SMDM) \cite{nie2024scalingmaskeddiffusionmodels} and DiffuLlama \cite{gong2024scalingdiffusionlanguagemodels}. These works execute interesting approaches to for scaling MLM models for discrete diffusion. SMDM analyzes MLM scaling laws and is trained in a compute-optimal manner. DiffuLlama distills an MLM model from an existing AR model, namely LLaMA 7B \cite{touvron2023llama}. While these approaches are viable and worthwhile, in the following experiments we shall argue that MARIA is the most pragmatic approach for scaling masked infilling models.
\section*{Downstream Perplexity}
Generative models optimize maximum likelihood objectives, and a common way to compare modeling performance is with likelihood on a test set. Here, we compare a similar notion of perplexity, which is defined as the exponentiated average negative log likelihood on some corpus of tokens. We select five standard datasets to evaluate downstream perplexity: WikiText \cite{merity2016pointer}, LM1B \cite{chelba2014billion}, Lambada \cite{paperno2016lambadadatasetwordprediction}, AG News \cite{zhang2016characterlevelconvolutionalnetworkstext}, and ArXiv papers \cite{clement2019arxiv}. Some of the datasets are tokenized for an MLM word level tokenizer, so we detokenize them following standard procedure \cite{sahoo2024simpleeffectivemaskeddiffusion}. Because the context lengths of models differ, we also compute fixed length perplexity on a rolling basis, that is partitioning corpuses of tokens as necessary to fit within a context and summing over the negative log likelihoods for each partition. We compute the perplexity given 5 different masking rates: 0.1, 0.3, 0.5, 0.7, 0.9 (least to most masked); specifically, the goal is to model the randomly masked tokens given the surrounding unmasked context. From the downstream datasets, we subsample 500 examples from each.

Importantly, discrete diffusion models do not admit an exact perplexity. Instead, we compute the negative evidence lower bound (NELBO) though sampling. While it may seem unintuitive to compare exact perplexities with upper bounds, in practice these bounds are tight \cite{kingma2023variationaldiffusionmodels}, and these comparisons are widespread in the literature \cite{ho2020denoisingdiffusionprobabilisticmodels, gulrajani2023likelihoodbaseddiffusionlanguagemodels}.

We report the average perplexities for seven models. ModernBERT perplexity is computed using the left to right autoregressive distribution that an MLM model admits by successively unmasking and computing the likelihood from left to right. We also compute the perplexities for regular AR models that cannot condition on future tokens. These results show that MLM models poorly model heavily noised text. We speculate that for ModernBERT, which was trained at a fixed mask ratio of 0.3 \cite{warner2024smarter}, performs poorly with higher noise ratios because they are out of distribution. Meanwhile, AR models cannot condition on future context and therefore demonstrate surprisingly strong performance independent of noising rate. MARIA, which is a mixture of OLMo and ModernBERT, achieves the upside of both models with strong performance in low noise settings, and it stays strong as the noise level increases, similar to the AR models. Of note, performance scales with model size, indicating a straightforward way to scale masked infilling capabilities more efficiently than scaling MLM models.

\section*{Throughput}
\begin{figure}[t]
    \centering
    \includegraphics[width=1\linewidth]{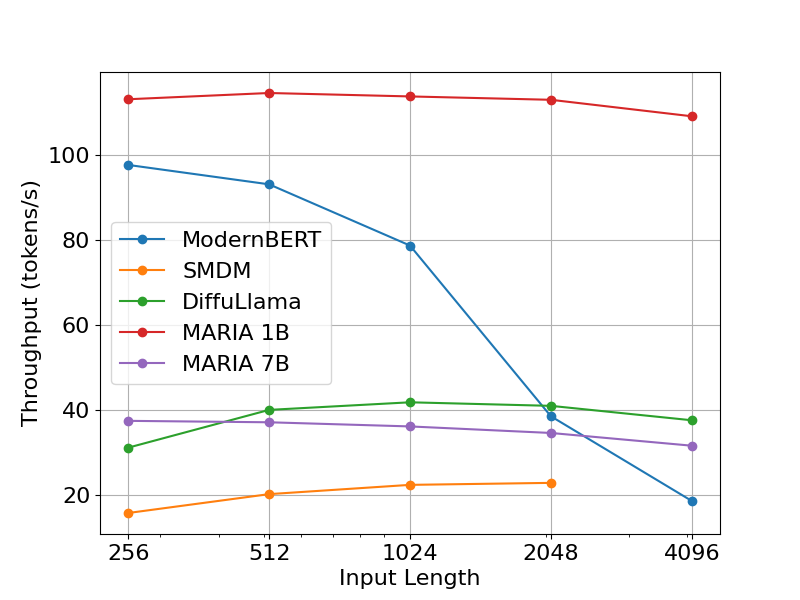}
    \caption{\textbf{Throughput over input length.} We show the throughput in tokens per second for sequences of given lengths at 0.5 masking rate. MARIA 1B exhibits the best performance, and MARIA-7B is comparable to DiffuLlama 7B. Decoding ModernBERT autoregressively is extremely inefficient at scale, and therefore is impractical in many circumstances.}
    \label{fig:throughput}
\end{figure}

Efficiency is a crucial reason why AR models are more widely adopted than MLM models. We profile the throughput of each model to better understand how these approaches compare. In light of this, we fix the generation parameters such as number of diffusion steps to the same parameters that will be later used in infilling experiments. Thus, we can analyze these efficiency results with sample quality results in tandem. In Figure \ref{fig:throughput}, we measure the throughput in tokens per second on different length inputs with 50 percent masking. We average the throughput over 10 runs, with 2 warm-up runs in the beginning for each model to ensure the GPU is operating maximally. From the results, we observe that MARIA 1B has the best throughput. Surprisingly, SMDM has worse throughput than larger 7B models. This can be attributed to an expensive classifier free guidance method (which we apply for later results) and miscellaneous implementation details. From Figure \ref{fig:throughput}, it is also critical to observe the performance of ModernBERT. Because ModernBERT is an MLM model incapable of KV caching, it is impractical to use for inference. KV caching models will have an inference runtime $O(n^2)$ in the sequence length, and without caching this runtime is $O(n^3)$. Though we include decoding ModernBERT autoregressively in the experimental benchmarks, poor efficiency at scale makes it severely impractical. Though discrete diffusion models cannot KV cache, they can unmask multiple tokens at each iteration. Thus, we see that DiffuLlama and MARIA 7B have similar throughputs. However, we shall show in the following section that MARIA achieves much better quality for similar efficiency.
\section*{Sample Quality}
To evaluate sample quality, we adopted the same setting as before using 1000 samples total from the downstream datasets previously described (200 samples each). The task is to infill a random 50 percent of the text for each. However, to ensure comparable masked sequences in light of different tokenizers, we mask 50 percent words by replacing them with the mask string (i.e. \texttt{[MASK]}), ensuring that every model is given the same task. We define a word to be an alphanumeric string with spaces at the beginning and end.

We set the inference time hyperparameters to the respective values that achieved the best results for DiffuLlama and SMDM. For DiffuLlama, it uses nucleus sampling \cite{holtzman2020curiouscaseneuraltext} and temperature scaling of 0.9 each. For SMDM, it applies classifier guidance scaling of 2 with greedy sampling. In all of the following experiments, we use 256 denoising steps. For ModernBERT and MARIA models, we decode greedily.

\begin{figure}[t]
    \centering
    \includegraphics[width=0.82\linewidth]{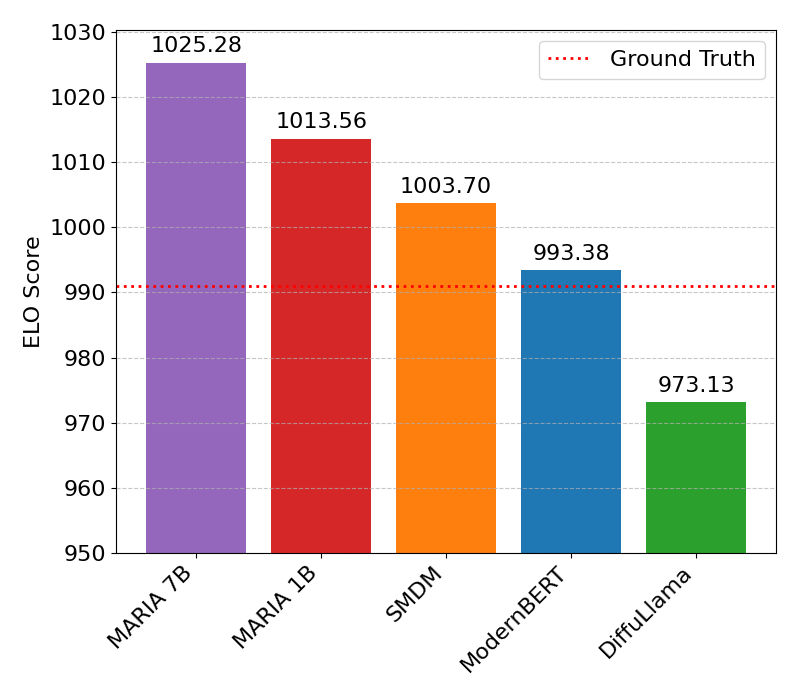}
    \caption{\textbf{ELO scores for masked infilling.} We perform infilling on downstream data with words masked 50 percent. Using GPT4o-mini as a judge we compute the ELO scores for each model respectively. MARIA 7B and 1B have the highest rating ELO rating under the Bradley-Terry model.}
    \label{fig:elo}
\end{figure}

We assess sample quality using an ELO system judged by GPT-4o mini \cite{achiam2023gpt}. We create 1000 random fixtures and prompt GPT to give a score for each text ``based on coherence, fluency, and style". For ELO scoring, a higher score is a win (1), lower score is a loss (0), and even score is a tie (0.5). We then calculate the ELO through logistic regression using the Bradley-Terry model, the same method as ChatBot Arena \cite{chiang2024chatbotarenaopenplatform}. This method ensures that match order does not influence the final score, which is a problem with iteratively computing online ELO. We employ standard hyperparameters of scale 400, base 10, and inital rating of 1000.

As shown in Figure \ref{fig:elo}, the MARIA models score the highest ELO ratings, with MARIA 7B and 1B attaining the top scores. In the ELO rating system, every difference of 400 corresponds to a 10x improvement in winning odds. From these results, we infer that the win probability of MARIA 7B against SMDM and DiffuLlama are 53.1\% and 57.4\%. Though these differences are not drastic, in practice it is difficult to achieve large differences in win rate if the LLM judge is insufficient to adequately differentiate between texts. Interestingly, the LLM judges the generated texts of four models as higher quality than the ground truth unnoised text. This may be a consequence of greedy decoding producing more likely text than the source text.

\section*{Test Time Scaling}

\begin{figure}[t]
    \centering
    \includegraphics[width=1\linewidth]{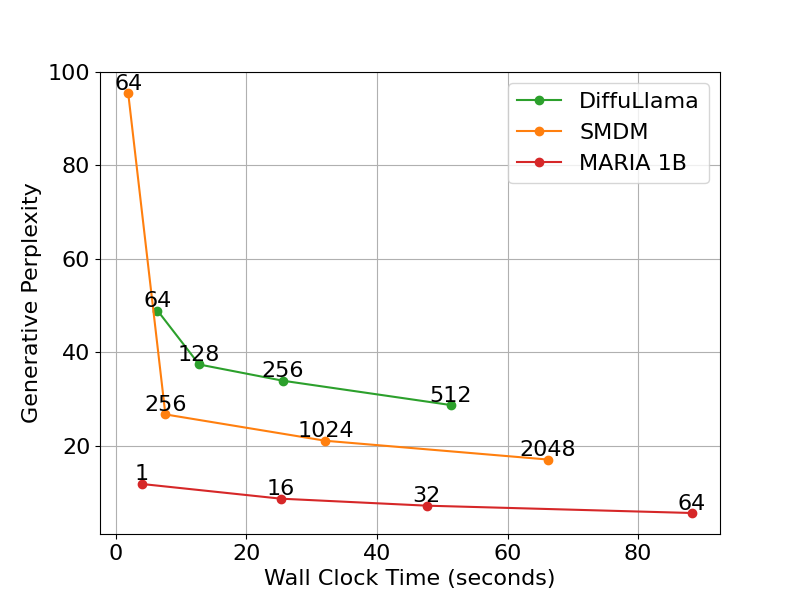}
    \caption{\textbf{Scaling test time compute for unconditional generation.} We compare our simulated annealing inference approach for MARIA to our baseline discrete diffusion methods. MARIA 1B using simulated annealing effectively trades-off quality (as measured by generative perplexity) and with compute (measured in wall clock time).}
    \label{fig:inference_scale}
\end{figure}

Discrete diffusion admits a desirable property that more FLOPs can be spent at test time to produce higher quality text. We discuss an alternative method for test time scaling in Section \ref{sec:method}, namely simulated annealing. We apply simulated annealing in MARIA by remasking 30 percent of tokens at each iteration and sampling with MARIA with a progressively lower temperature using a linear schedule. In Figure \ref{fig:inference_scale}, we measure the generative perplexity of 200 unconditional samples according to Llama3 8B \cite{grattafiori2024llama3herdmodels}. We show that for MARIA 1B, simulated annealing is an effective and efficient way to generate higher quality samples, converging faster than both DiffuLlama and SMDM. MARIA 7B with simulated annealing is far slower to converge than MARIA 1B, and it is omitted to avoid plot scaling issues.

\section*{Representations}

Representation learning is a key motivation behind training Transformers with an MLM objective. We aim to analyze MARIA through a representation learning perspective to offer insight into why combining MLM and AR models can improve performance. Specifically, we study the token level representations by measuring performance on part-of-speech tagging. The part-of-speech tagging task has a history in NLP \cite{manning2011part}, and we use the CoNLL-2003 dataset \cite{sang2003introductionconll2003sharedtask}. We train a linear classifier on representations from ModernBERT, MARIA 1B, and MARIA 7B on 10000 sentence examples with POS labels that can belong to 48 different classes. We train for 10 epochs with a learning rate of 1e-4. As Table \ref{tab:representation} shows, part-of-speech tagging accuracy increases with MARIA 1B and further increases with MARIA 7B. These results are somewhat expected because MARIA hidden states are much larger in dimension: ModernBERT has dimension 1024, MARIA 1B has dimension 3072, and MARIA 7B has dimension 5120. These results confirm that AR representations contain information that MLM representations do not due to scale.

\begin{table}[t]
\centering
\vspace{0.5cm}
\begin{tabular}{l l}
\toprule
Representation & Accuracy  \\
\midrule
ModernBERT & $0.642 \pm 0.002$ \\
MARIA 1B & $0.714 \pm 0.002$ \\
MARIA 7B & $0.735 \pm 0.002$ \\
\bottomrule

\end{tabular}
\vspace{0.5cm}
\caption{\textbf{Representation learning for part-of-speech tagging.} We demonstrate that MARIA representations produce higher accuracy when used to predict parts-of-speech. This indicates that the concatenated AR and MLM hidden states of MARIA contain more information than MLM alone.}
\label{tab:representation}

\end{table}

\section{Conclusion}
The introduction of MARIA (Masked and Autoregressive Infilling Architecture) addresses a long-standing gap in the field of natural language processing by seamlessly combining the strengths of autoregressive (AR) and masked language models (MLM). This hybrid approach has demonstrated significant improvements in masked token infilling, achieving lower perplexity scores across diverse datasets and outperforming existing methods like discrete diffusion models in both quality and efficiency. Furthermore, MARIA's integration of KV caching ensures it retains the computational advantages of AR models during inference. 

Future directions include further optimizing the inference algorithm to support modern AR inference techniques. For example, incorporating Paged Attention \cite{kwon2023efficient} would provide tremendous gains in throughput beyond the gains demonstrated in this paper. Also, in this paper, we utilize a pretrained base AR and MLM model. For future work, it is possible to use fine-tuned versions of these models for domain-specific tasks. For example, combining an AR and MLM model specialized for infilling DNA sequences or code blocks could yield strong, highly specialized infilling models.

\section*{Impact Statement}
This paper presents work whose goal is to advance the field of 
Machine Learning. There are many potential societal consequences 
of our work, none which we feel must be specifically highlighted here.

\newpage
\bibliography{references}
\bibliographystyle{icml2025}

\end{document}